# Machine Learning in the Social and Health Sciences

Anja K. Leist[1*], Matthias Klee[1], Jung Hyun Kim[1], David H. Rehkopf[2], Stéphane P. A. Bordas[3], Graciela Muniz-Terrera[4], Sara Wade[5]

**Affiliations**
[1]University of Luxembourg, Department of Social Sciences – Institute for Research on Socio-Economic Inequality (IRSEI), Esch-sur-Alzette, LU
[2]Stanford University, Department of Epidemiology and Population Health, Palo Alto, CA
[3]University of Luxembourg, Department of Engineering, Esch-sur-Alzette, LU
[4]University of Edinburgh, Centre for Dementia Prevention, Edinburgh, UK
[5]University of Edinburgh, School of Mathematics, Edinburgh, UK

*Corresponding Author: Anja K. Leist, Associate Professor, University of Luxembourg, Department of Social Sciences – Institute for Research on Socio-Economic Inequality (IRSEI), Campus Belval, 11, Porte des Sciences, L-4366 Esch-sur-Alzette, Luxembourg. Phone: +352 46 66 44 9581, anja.leist@uni.lu

**Acknowledgements**
We would like to thank Dr Benedikt Wilbertz, Dr Zhalama, Thiago Quaresma Brant Chaves, as well as the DEMON (demondementia.com) Prevention Working Group and our colleagues at IRSEI for helpful discussions.

**Funding**
This work was supported by the European Research Council (grant agreement no. 803239, to AKL).

**Conflict of Interest Statement**
The authors declare no conflict of interest.

## Abstract

The uptake of machine learning (ML) approaches in the social and health sciences has been rather slow, and research using ML for social and health research questions remains fragmented. This may be due to the separate development of research in the computational/data versus social and health sciences as well as a lack of accessible overviews and adequate training in ML techniques for non data science researchers. This paper provides a meta-mapping of research questions in the social and health sciences to appropriate ML approaches, by incorporating the necessary requirements to statistical analysis in these disciplines. We map the established classification into description, prediction, and causal inference to common research goals, such as estimating prevalence of adverse health or social outcomes, predicting the risk of an event, and identifying risk factors or causes of adverse outcomes. This meta-mapping aims at overcoming disciplinary barriers and starting a fluid dialogue between researchers from the social and health sciences and methodologically trained researchers. Such mapping may also help to fully exploit the benefits of ML while considering domain-specific aspects relevant to the social and health sciences, and hopefully contribute to the acceleration of the uptake of ML applications to advance both basic and applied social and health sciences research.

## Significance statement

There is great interest in the social and health sciences in the application of machine learning (ML) methods, however, a conceptual mapping of appropriate ML approaches to research questions in the social and health sciences has been lacking. The classification presented here may help to advance the uptake of ML in social and health sciences while also pointing to possible limitations and ways of addressing them.

## Introduction

Compared to many traditional statistical methods, in particular with increasing availability of large datasets of relevance to the social and health sciences, machine learning (ML) methods have the potential to considerably improve aspects of empirical analysis. This includes advances in prediction, by fast processing of large amounts of data, in detecting non-linear and higher-order relationships between exposures and confounders, and in improving accuracy of prediction. However, uptake of ML approaches in social and health research, spanning from sociology, psychology and economics, to social and clinical epidemiology and public health, has been rather slow and remains fragmented to this date.

We argue that this is in part due to a lack of communication between the disciplines, the importance of incorporating domain knowledge into statistical analysis in the social and health sciences, and a lack of accessible overviews of ML approaches fitting the research goals in the social and health sciences. Further, computational fields have traditionally emphasized improvements in prediction, whereas the social and health sciences have often prioritized explanation (Yarkoni & Westfall, 2017) and the importance of domain knowledge. There is a need to establish a fluid dialogue between researchers from the social and health sciences and methodologically trained researchers to avoid "rediscovering the wheel". ML researchers may overlook features of data that have been previously found to be highly relevant in the social and health sciences (e.g., oversimplification of recoding of some variables when integrating datasets). In contrast, researchers in the social and health sciences may not be aware of the complex mathematics and statistics behind the algorithms and the fast progressing developments of improving ML methods, and how the mathematical commonality between different domain problems can be leveraged to improve the efficiency of data-driven research (Ley & Bordas, 2018).

The aim of this paper is to provide a high-level, non-technical toolbox of ML approaches through the systematic mapping of research goals in the social and health sciences to appropriate ML methods. These can be built on top of each other or combined with more traditional methods. We will further point researchers to solutions to common problems in ML modelling. As we have strong interests in the social and behavioural determinants of health and disease, many applications mentioned here focus on relevant research and possible ML approaches in the fields of psychology, social epidemiology, and prevention research.

This conceptual overview should be seen as complementary to several introduction papers to machine learning in the fields of epidemiology and health research (Wiemken & Kelley, 2020), psychology (Yarkoni & Westfall 2017), and helpful glossaries (Bi et al., 2019). For general introductions to statistical learning, interested readers are referred to several excellent textbooks on these approaches (Efron & Hastie, 2016; J. Friedman et al., 2001; Hardt & Recht, 2021; James et al., 2013). The changed research infrastructure and computational requirements when using machine learning and issues of privacy have been discussed elsewhere (Fuller et al., 2017; Mooney & Pejaver, 2018) and will not be covered here.

While some of the ML approaches presented here require more domain knowledge than others, for example ML for causal inference, we argue that in all research questions in the social and health sciences substantial domain knowledge is necessary to meaningfully contribute to the field, and that this is a prerequisite to interpretability (Murdoch et al., 2019). Agnostic data exploration alone will in most cases provide fewer insights. Generally, from our own experience, we recommend collaborations across disciplines by inviting data science and machine learning experts to do research in the social and health sciences, and hope the mapping presented here will facilitate mutual understanding of the different disciplines.

We present here, adapted to the social and health sciences, a classification of machine learning tasks for description, prediction, and causal inference (Hernan, Hsu & Healy 2018), even if not all research questions allow such strict distinctions. We will illustrate this mapping with studies from the social and health sciences. The methods summarized as ML in this overview represent different traditions of data analysis, e.g. inferential statistics, statistical learning, and computational sciences, their common denominator is the ability to process large amounts of data while making model-building and model selection decisions more driven by the structure of the data (data-driven) than traditional inferential statistics. First, we provide a set of basic terms and the ML workflow and general points on interpretability, fairness, and generalizability, before describing the ML tasks in more detail.

## Background to ML approaches

ML approaches will usually involve the 'training' (estimation) of a model in a so-called training data set and, in a second step, the 'testing' of the model regarding its performance (e.g., accuracy of classification) in a separate test data set. ML approaches can be categorized into unsupervised learning, supervised learning, and reinforcement learning:

- Unsupervised learning is an umbrella term for algorithms that learn patterns from unlabelled data, that is, variables that are not tagged by a human. For instance, unsupervised learning will group data instances based on similarity.
- Supervised learning comprises algorithms that learn a function which maps an input to an output, by using labelled data, that is, the values of the categories of the outcome variable are assigned meaningful tags or labels. *Input* would in the social and health sciences be termed predictors, independent variables or exposures, and covariates; *output* would be termed outcomes or dependent variables. Supervised learning requires labelled training data and can be validated in a labelled test data set. We will present neural networks (often called artificial neural networks, ANNs) as an example of a predictive algorithm (Box 1), and Bayesian Additive Regression Trees (BART) as an example of ML for causal inference (Box 2). *Transfer learning* will transfer learnt features from one situation to another (congruent) situation, thereby identifying patterns and behaviours common to a variety of situations. While often employed with labelled data and thus mentioned as approach specific to supervised ML, transfer learning approaches have also been developed for image recognition and other applications not covered in this review.
- Reinforcement learning is concerned with an intelligent agent taking decisions to an environment and improves based on the notion of cumulative reward, i.e. the agent will vary and optimize the input based on the feedback from the environment. Reinforcement learning can be applied in contexts where data generation, that is, manipulation of a treatment variable (so called A/B testing) under control of other features (covariates) is possible.

Most ML approaches presented here will require large(r) datasets than traditional modelling for the models to outperform traditional modelling in new data sets. Some aspects of the data should be available at a larger quantity, such as time points, variables, or individuals. Rule of thumb would be to have several ten thousand data points available, but some applications have used very small datasets for exploratory analyses (Wiemken & Kelley 2020). Similar to non-ML analyses, careful processing of data (e.g., during the harmonization process), a deep understanding of where the data are coming from, what they can tell us (and what they cannot), is vital. The ML workflow has been described in several introductory papers, see Wiemken & Kelley (2020) for a recent overview.

In this paper, we will not cover ML algorithms that have been developed for more automatic processing of big data, such as speech recognition and sentiment analysis. These algorithms have been specifically developed to increase speed and efficiency of analysis of speech, text or images. Use of ML in these fields indeed provide numerous advantages over traditional procedures in qualitative social research such as higher efficiency and accuracy in retrieving summary features. Other use cases of ML concern the anonymisation of electronic health records to ensure GDPR regulations prior to analysing sensitive data or the preparation of infrastructures to analyse large datasets most efficiently (e.g. reinforcement learning to optimize batching of data to be analysed). Our review focuses on research questions that involve datasets with human participants as research units and the analysis of clinically assessed or self-reported variables.

In the following, we present a few additional aspects of particular relevance to the social and health sciences. In the *data preparation process*, researchers need to prepare exposure and outcome variables (so-called feature selection and engineering) in a way the algorithm understands: *Feature selection*, that is, reducing the number of variables to be processed by the model by applying the minimal redundancy maximal relevance criterion, may be helpful in large datasets such as the PISA survey or aging surveys from the family of Health and Retirement Studies. Conceptually interesting features, for example cumulative risk (multiplicative effect of two predictors) or changes between measurements (e.g. weight loss over time) are not well detectable by including the single variables. We suggest to explore if the reduction of complexity in the set of (related) independent variables makes sense, e.g. factor analysis or cluster analysis, and/or selection of variables based on theory. If the dataset is large enough, a rule of thumb here is the availability of several 10,000 relevant units (e.g. respondents to a survey), this enables the use of neural networks, which are well known to use feature engineering in the generation of the models. While it is necessary to make the continuous variables equivalent in variance, we would recommend using manual feature engineering only to an extent to which researchers in the social and health sciences can still ensure some interpretability for real-world applications. On the other hand, traditional modelling is highly depending on researcher decisions (e.g., modelling a quadratic instead of a linear relationship, modelling interaction effects manually). Here, algorithm-based decisions regarding feature engineering for example in neural networks can provide more robust and accurate findings. Approaches like BART (described in more detail below) will handle the simultaneous inclusion of continuous, dichotomous and categorical predictors. In contrast, other approaches such as regression trees will overvalue continuous predictors simply due to the availability of a larger number of possible splitting points, however rules for splitting decisions for categorical variables exist in some algorithms, but are handled differently across software packages.

In feature engineering, researchers should always be conscious of the *curse of dimensionality,* which describes the tendency of the test error to increase as the dimensionality of the problem increases, unless additional features are truly associated with the response (i.e., not just adding noise). More features, that is, variables in the model, increase the dimensionality of the problem, exacerbating the risk of overfitting. Thus, advances in data acquisition that allow for the collection of thousands or even millions of features are a double-edged sword; improved prediction can result if the features are truly relevant and the sample is population-representative, but they will lead to more biased results if not. Moreover, even if they are relevant, the reduction in bias may be outweighed by increased variance incurred by their fit (James et al., 2013). Evaluating the *quality and performance of the model*, ML models offer less straightforward solutions compared to more traditional modelling in the social and health sciences. To overcome this, we have compiled a table with an overview of model performance indicators for each of the ML approaches (table 2). Particularly ML for regression and classification can also be compared against traditional methods to see the value gained.

Researchers also need to be aware of the *trade-off between increasing predictive accuracy and overfitting*, ideally through proper cross-validation procedure. Most researchers will be familiar with the ML modelling process which comprises splitting the data into a training and a test set, possibly also a third partition which is held-out of the train-test iterations for later model evaluation. External cross validation further allows testing the generalization error of resulting models. In newer packages such as the SuperLearner (a.k.a. stacking) that wraps a number of different algorithms to increase model performance, solutions are in-built already. It is preferred to additionally validate models in new datasets (neither used for training nor test). We illustrate this trade-off between predictive accuracy and overfitting in Figure 1, which highlights the typical interrelation of training, validation, and test errors. The training error typically decreases with model complexity, while the validation and test curves have a U-shape. Underfitting refers to the case when a more complex model improves the test error, and overfitting occurs when a less complex model yields a lower test error. We almost always expect the training curve to lie below the test curve, as most methods aim to minimize the training error. Importantly, the test data is held-out in model training in order to provide a realistic measure of performance, thus validation methods, which split the training dataset, are employed to select complexity and tuning parameters. The validation curve typically lies above the test curve, since it is trained on a smaller training set, and can also be highly variable due to the random data splits. However, the goal of validation is to identify the correct level of flexibility, i.e. the minimum point of the test error. Cross-validation is typically used to help reduce the variability in the validation curve. However, it is appropriate under the assumption of independent and identically distributed data; for some data, such as time series or longitudinal data, this is not appropriate and splits must account for the structure in the data. When the validation curve is relatively flat, the simpler model is preferred.

AutoML approaches such as the H2O.ai interface aim to facilitate the ML workflow for non data science researchers (H2O.ai, 2021; LeDell & Poirier, 2020). AutoML by default trains and cross-validates generalized linear models, gradient-boosting machines, random forests, deep neural networks and combines via SuperLearning/stacking to improve model fit. While easy-to-use interfaces are attractive to non data sciences researchers and can be helpful in many cases, we argue that, similar to research with traditional inferential statistics, an understanding of the applied methods in greater detail is still necessary for substantial research contributions.

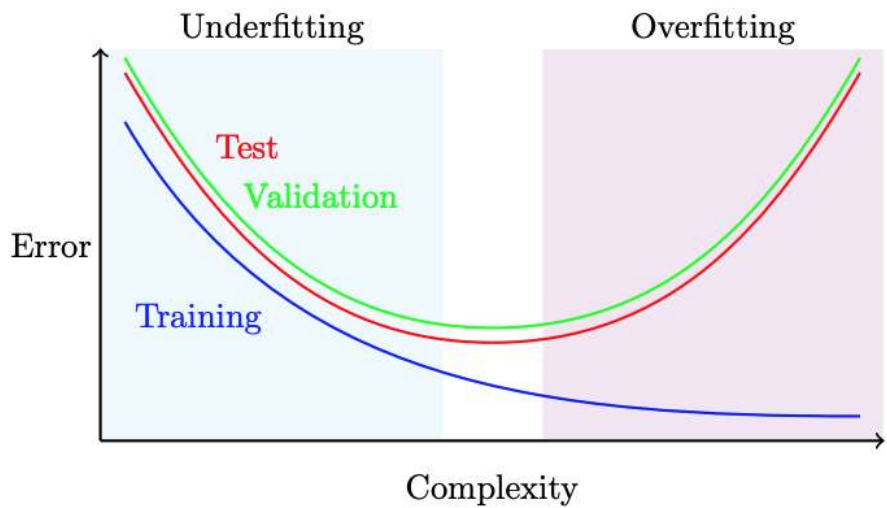

*Figure 1. Typical relationship between model error and complexity.*

## General points

In the following, we will refer to relevant concepts when applying ML in the social and health sciences, specifically, interpretability, fairness, and generalizability.

*Interpretability / explainability and visualization*
Interpretable ML means to extract relevant knowledge from ML models, i.e. able to feed into the domain knowledge of a discipline, characterized by predictive accuracy, descriptive accuracy and relevancy, with relevancy judged relative to a human audience (Murdoch et al., 2019). Interpretability is important in high-stake decisions, such as clinical decision making. Here new developments exist to increase acceptance, e.g. the symbolic metamodeling framework for interpreting predictions by converting "black-box" models into "white-box" functions that are understandable to humans (Alaa & van der Schaar, 2019). Others have however suggested that the use of separate, explainable, models could be problematic in applications as there is still limited transparency for the users of the ML models (Rudin, 2019).

Interpretability is in practice often linked with the possibility to visualize the estimated or discovered relationships among variables. Summary packages intended to give a user-friendly set of possibilities to increase (post-hoc) interpretability have been developed, e.g. iml (Molnar et al., 2018). To increase interpretability, bi-variate and higher-order data associations with possible non-linear patterns or missing data that are not at random, as well predictions can be visualized e.g. with partial dependence plots (Greenwell, 2017), SHAP plots (Lundberg et al., 2018; Mangalathu et al., 2020), or Individual conditional expectation (ICE) plots, a newer and less well-known adaptation of partial dependence plots (P. Hall et al., 2017).

*Fairness*
A relevant notion in the health and social sciences is the idea of the ML algorithm to decide "fairly", i.e. will not discriminate against certain social or minority groups. Here, particularly in the application of ML in high-stake decision making such as predicting recidivism, predictive model accuracy needs to be balanced against the model treating all social groups equally, i.e. the ML model resulting in all social groups having equal probability of receiving a desirable outcome. We refer to other literature for considerations regarding fairness in ML at large (Barocas et al., 2017) and ethics in ML in the human and social sciences (Lo Piano, 2020). Importantly, health equity efforts can be undermined by structural discrimination/racism possibly implemented in ML diagnosis or decision-making (Rajkomar et al., 2018; Robinson et al., 2020). Outside the scope of our review, it is critical to realize that fairness is not just important in application of algorithms, but in the full arc of the research process (Chen et al., 2020).

*Generalizability or external validity*
The importance of generalizability of external validity is not limited to ML approaches: Across research designs as well as across stages of data collection, we need to consider biases that could harm generalizability, as it is desired that findings have validity beyond the dataset in which they were discovered. In the context of ML, it is important to mention the risk of overfitting, i.e. to improve model accuracy within the dataset by risking model performance in a new dataset. An example of how differences in recruitment in cohort studies can result in differential performance of ML algorithms is (Birkenbihl et al., 2020). And, similar to more traditional analyses, triangulation of methods is recommended (Matthay et al., 2020). While in the past it has been stated that a lot of 'black art' necessary to successfully implement ML in research (Domingos, 2012), we stress the importance of documenting all decisions in the statistical analysis to the details (e.g. specifying seed of random number) to ensure scientific findings are replicable.

We now move to the classification of ML approaches for description, prediction, and causal inference (some of which admittedly have their roots in standard statistical methods), and start with ML approaches for description.

## ML for Description

A descriptive research question aims to "provide a quantitative summary of certain features of the world" (Hernán et al., 2019). Description is the basis of all applied research, as we need description to quantify the phenomenon under study, e.g. assess prevalence or distributions of variables between (social) groups, countries or geographical entities, or over time (between cohorts). This can be done through several algorithms (Table 1). Among these, researchers in the social and health sciences are aware of factor analysis of one (cross-sectional) measurement of several variables. Dimensionality reduction can reduce complexity of datasets for more efficient subsequent analysis (van der Maaten et al., 2009). Domain knowledge is necessary to pre-select variables for factor analysis that are interpretable and meaningful, as cluster (and factor) analysis cannot conceptually distinguish variables, for example, if some data are coming from humans or from other data. Factor analysis will provide factors and factor loading for the included variables. Factors are often used as new variables with densified information in subsequent analyses. Readers should note that algorithms based on Bayesian modelling are available for factor analysis and other methods presented here but will not be covered in detail. We suggest to not mix individual-level variables with higher-order variables e.g. related to environment or neighbourhood in a factor analysis to ensure interpretability of the factors in subsequent analyses. Another helpful approach may be cluster analysis to group data based on similarity. Unlike in factor analyses, cluster analysis can help to process measurements of individual-level and contextual-level variables simultaneously, for example, individual-level BMI and contextual-level air pollution to investigate different risk groups or profiles (e.g. with high BMI and high air pollution) for dementia. Finally, bi-clustering may be helpful if samples and variables need simultaneous grouping.

**Screen and identify individuals at risk or higher-level patterns**
Individuals can be screened for *single risk factors*, through automated processing of data, for example, people at elevated risk for adverse health outcomes can be identified through processing of Electronic Health Records. This research goal could be solved with algorithms for anomaly detection. Conducting clustering or factor analysis, a *group of variables* can be processed simultaneously, by analysing patterns across individuals. To identify ageing-related morbidity pathways, electronic health records that contained information on 278 high-burden diseases were analysed with different clustering algorithms to group diseases according to their patterns of age at onset of disease (Kuan et al., 2021). To add, the complex information derived from ML was visualized with traditional plotting, for example, distributions of onset of disease curves per disease cluster (Kuan et al., 2021).

Longitudinal analyses of *trajectories of time-varying variables* can be helpful to better understand the trajectory of previously identified risk factors with long-term observational data, preferably with minimum 5 follow-up measurements. This can be interesting in diseases where long-term risk prediction is relevant, or where long prodromal phases need to distinguish if factors are indeed risk factors or early symptoms. Depressive trajectories were identified via k-means clustering of the number of depressive symptoms at each measurement occasion over a long follow-up, to distinguish early-onset from late-onset depression (Demnitz et al., 2020). Analyses like this can help elucidate changing risk factor importance over the life course, for example in the field of dementia (Peters et al., 2020)

Death as a competing risk needs to be accounted for in investigations on aging-associated diseases, e.g. with random survival forests, which also allow modelling of time-varying risk factors (Weiss et al., 2020). If a focus is on the short-term consequences of time-varying treatment, i.e. if causal conclusions are intended, it may be better to use established methods in the potential-outcomes framework such as marginal structural models or the g-formula (Robins et al., 2000; Hernán & Robins 2021).

**Risk profiles**
To identify risk profiles, i.e. groups of individuals characterized by certain values on a set of variables, ML for discovery can be used to describe and reduce complexity, based on previous literature that had identified risk or protective factors. Using severity scores common in the ICU setting, patient health state trajectories were categorized with a number of dimensionality reduction (and predictive) techniques in time-series data, among others DBSCAN (Galozy, 2018). These health state trajectories were correlated with medication and treatments, with commendable visualizations of the resulting patterns (Galozy, 2018).

While most descriptive research problems in the social and health sciences will require unsupervised ML, there is no one-to-one correspondence of descriptive problems to unsupervised ML. In the following, we will present descriptive research questions that need supervised ML approaches typically used for prediction.

**Diagnosis**
A descriptive research goal is to identify prevalence of a health outcome, which can be reasonably inferred also in absence of clinical assessment. This descriptive goal of diagnosis can be accomplished with supervised ML or a semi-supervised setting if a mix of labelled and unlabelled data is analysed. With data from electronic health records (EHR), algorithms can, in the absence of human clinical assessment, identify the existence of characteristics (or joint presence of conditions) that increase likelihood of a presence of disease, for example through neural networks. Other data, such as sensor data, language data etc. can be used to detect conditions or disease. Movement of sign language users was analysed to detect likely dementia with convolutional neural networks (Liang et al., 2020).

In the absence of a diagnosis based on clinical assessments, classifying individuals with a probable diagnosis through ML may be interesting to *estimate population-level disease prevalence* and associated healthcare costs. Identifying individuals with probable diagnosis is unobtrusive and may be more cost-effective than the clinical assessment of these individuals. Combining PCA and cluster analysis, participants with high likelihood of dementia were identified in U.S. and European datasets (de Langavant et al., 2018) and with datasets from across the world (de Langavant et al., 2020). With cultural and education fair battery of cognitive tests, the 10/66 diagnosis of dementia can be done with cognitive tests alone to improve prevalence of dementia estimations in the absence of possibilities for clinical assessment, with application of machine learning to data from South India (Bhagyashree et al., 2018).

Supplementing classifications of probable diagnosis with diagnosed individuals may address *underreporting* and *underdiagnosis* of conditions such as dementia. These individual-level probable diagnoses can be used to investigate risk and protective factors of these conditions. Survey participants were identified as having probable dementia with a mix of traditional and ML based (descriptive and predictive) algorithms, and with the aim to provide dementia classification algorithms with similar sensitivity/specificity across racial/ethnic groups (Gianattasio et al., 2020). ML in this study proved more complex to implement and was considered more sensitive to cohort and study procedural differences than traditional modelling; additionally, use of ML, in this case LASSO and the SuperLearner, did not lead to increases in model performance compared to different expert models (Gianattasio et al., 2020). Samples of less than 2,000 participants may thus be less recommended for the use of ML approaches.

**Table 1. Overview and non-technical description of ML methods for description most relevant in the social and health sciences**

| ML for Description |
| --- |
| Clustering: |
| Clustering groups data based on similarity. Examples are |
| <ul><li>k-means: clusters data points according to a distance metric in an n(variables)-dimensional space</li><li>Hierarchical agglomerative clustering: starting from each object forming a separate cluster, clusters are consecutively merged moving up the hierarchy</li><li>Model-based clustering: the most widely-used example is the Gaussian mixture model (GMM), which generalizes k-means by allowing elliptically-shaped cluster.</li><li>Density-based spatial clustering of applications with noise (DBSCAN) (Ester et al., 1996; Schubert et al., 2017) groups data points in spatial proximity while marking data points in low-density regions as outliers.</li><li>Mixtures of experts (learners): This approach divides the input space in homogeneous regions, and a different expert is responsible for each region. This allows for different clusters, e.g. of patients, with different (non)linear relationships between y and x (Jordan & Jacobs, 1994; Masoudnia & Ebrahimpour, 2014).</li></ul> |
| Dimensionality reduction |
| <ul><li>Principal component analysis (PCA): Known to researchers in the social and health sciences, this method provides a low-dimensional approximation/encoding of the data by linear (orthogonal) projection (i.e. low-dimensional features are linear combinations of the original features)</li><li>Probabilistic principal component analysis (PPCA) is a probabilistic model formulation of PCA for higher-dimensional data such as found in metabolomics. The PPCA will provide the PCA solution in limit of zero noise (Tipping & Bishop, 1999)</li><li>Factor Analysis: Also known to researchers in the social and health sciences in the analysis of, for instance, questionnaire-based data, factor analysis can be seen as a</li></ul> |

generalization of PPCA that allows dimension-specific noise. This method explains the correlation across dimensions through a small number of latent factors.
- Independent Component Analysis: generalization of FA that allows the distribution of the latent factors to be any non-Gaussian distribution.
- Nonlinear dimension reduction: includes kernel PCA, Gaussian process latent variable model (GPLVM), t-SNE (Van der Maaten et al. 2009).
- Generative Adversarial Networks (GANs): The task of grouping data points based on similarity is split into a two-part problem of, first, generation of new data that should be similar to the real data, and the task of, similar to supervised learning, classifying the data as either real or new (fake). The task stops once the algorithm is no longer able to discriminate real from new data (Goodfellow et al., 2014).
- Variational Autoencoders (VAEs) employ neural networks for dimensionality reduction, both for encoding and decoding (mapping the data to the low-dimensional latent space and v.v.). VAEs use a probabilistic formulation and variational inference to learn the distribution of the latent variables, which avoids overfitting and imposes desirable properties on the latent space (Kingma & Welling, 2013)

Anomaly detection
This is the process of identifying data points that deviate from normal "behaviour", that is, is identified as dissimilar in the context of the overall data points. Anomalous data may indicate an incident, deviant behaviour (e.g. fraud in data of bank transfer, changes in household composition of a consumer in consumption data).

Biclustering
As a newer development, one may also be interested in the simultaneous grouping samples (individuals) and features (variables) based on similarity. The so-called biclustering methods simultaneously cluster samples and features (for a recent review: Padilha & Campello 2017). Biclustering is used in bioinformatics, e.g. to cluster patients based on expression profiles on a subset of genes (Moran et al. 2020).

## ML for Prediction

The ML task of *prediction* will need mapping some features (input) to other, known, features (output) as accurately as possible (Hernán et al., 2019). Again, known outcomes like a health outcome are called labelled data; to investigate predictive research questions, we thus employ **supervised learning**: As explained above, supervised learning is the ML task of learning a function from labelled data to map an "input" (predictors, independent variables) to an "output" (outcome). Numerical (continuous) outcomes will require *regression* techniques, while dichotomous or categorical outcomes will require *classification* techniques. Figure 2 presents an overview of ML methods that are, based on theoretical considerations, ranked to explain the trade-off between explainability versus complexity of these methods.

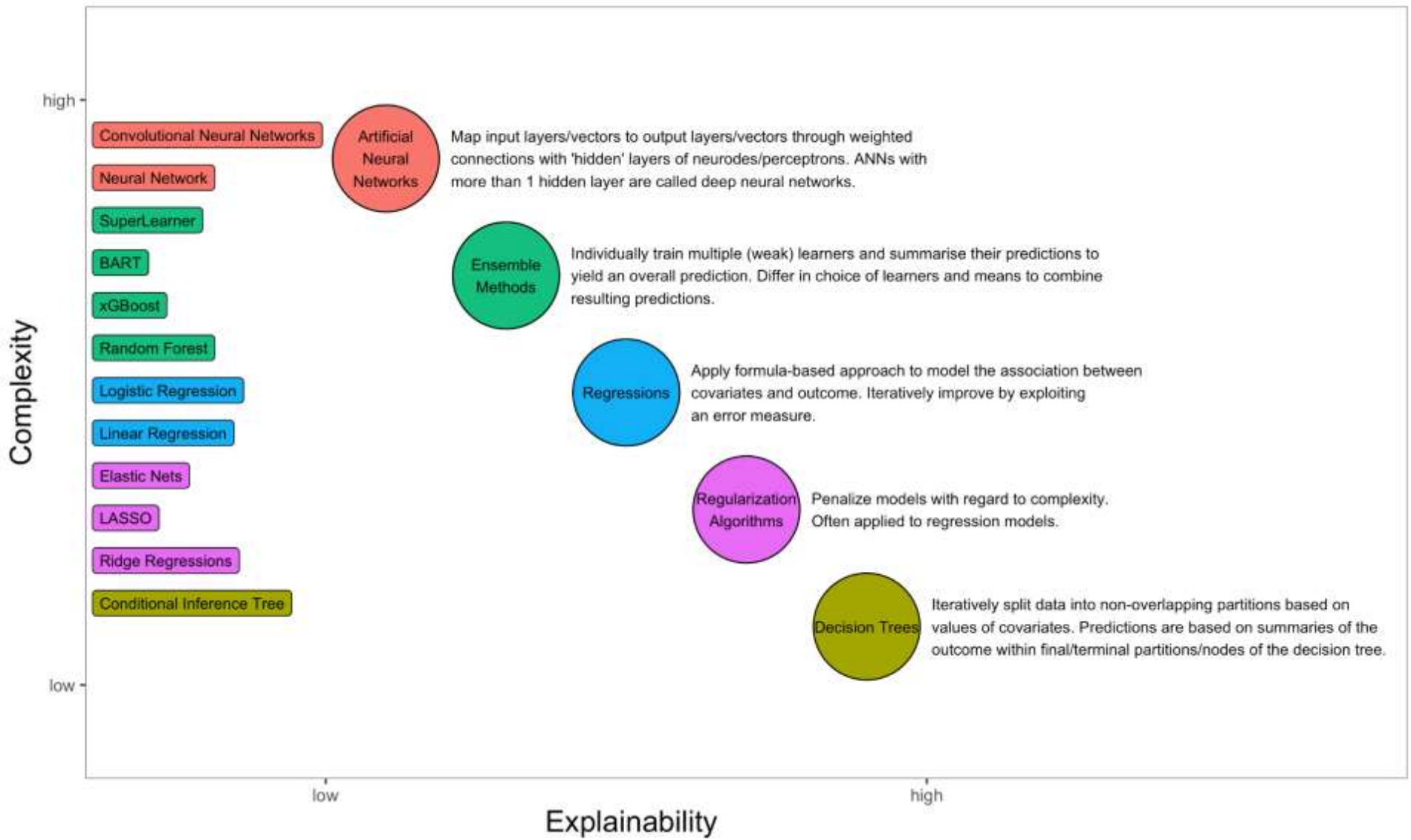

*Note.* Ordering and selection of ML methods based on theoretical considerations.

*Figure 2. ML methods for prediction most relevant in the social and health sciences with non-technical description ranked by interpretability/explainability versus complexity.*

As a few popular and/or well-performing ML methods, typical ML methods to solve predictive problems are penalized regression methods, ensemble learning, gradient boosting, decision trees, and neural networks (see Figure 2). Neural networks are described in more detail in Box 1. *Shrinkage or penalized regression methods* (e.g. LASSO, ridge regression, elastic nets) are also used in "traditional" inferential statistics. Penalized regression performs better than the standard linear model in large multivariate data sets where more variables than individuals are available. Penalized regression will add a constraint in the equation to penalize linear regression models with too many variables in the model, also called "shrinkage" or "regularization". This will shrink the coefficient values towards zero, so less contributive variables will have coefficients close to zero or equal zero (Hastie et al., 2009). *Ensemble learning* improves predictive accuracy by using multiple descriptive and predictive ML algorithms. An example family of algorithms is the SuperLearner that has been applied in epidemiological research questions (Naimi & Balzer, 2018). The SuperLearner uses cross-validation to estimate the performance of several descriptive and predictive ML models, or in the same model with different settings, and works asymptotically as accurately as the best prediction algorithm used in the model fitting process. Due to their interpretability, *decision tree algorithms* and ensemble methods such as Random Forests (Svetnik et al., 2003), have been employed in numerous studies in the social and health sciences. Commended for its high predictive accuracy and robust applicability to many predictive problems, we point readers to *stochastic gradient boosting* (J. H. Friedman, 2002), which has been implemented in ready-to-use packages in python and R for the family of Gradient Boosting algorithms, for example in the recommended xgboost package. Through boosting, the typical problem of collinearity of input variables (predictors) does not occur, that means that differently engineered (pre-processed) variables can be entered simultaneously to see which characteristics are most predictive. Researchers need to be aware that the algorithm does an exhaustive search over all variables for splitting-points, and some variables may be more informative to divide the sample than others. The algorithm is thus biased towards choosing numerical (continuous), multi-category variables or variables with missing data over dichotomous variables; methods towards unbiased variable selection are available (Hothorn et al., 2012). Further, variables will be picked as splitting points that best explain the dependent variable, which is not necessarily the most meaningful variable from a theoretical perspective relevant in the social and health sciences.

---

*Box 1: Detailed description: Artificial neural networks*

A more complex way of solving prediction problems by mapping some features (predictors) to other features (outcome) can be done with neural networks, which learn interrelationships between variables, with a defined "input" (predictors) and "result" (outcome). Neural networks are motivated by the computation in the brain that enables successful recognition and classification of complicated tasks (Fine 1996). Neural networks are identical to the traditional logistic regressions with no hidden layer if logistic activation function is implemented, which is the most common case (Dreiseitl and Ohno-Machado 2002). Both neural networks and logistic regression have a functional form, and the parameter vector is determined by maximum-likelihood estimation. However, neural networks allow us to relax the linearity of input variables and log odds assumption. Consequently, it is a better option if the data is not classified linearly. This flexibility comes with a cost of difficulty in interpretation of the parameters; the resulting model is evaluated through model performance measures such as sensitivity, specificity, accuracy, and the area under the ROC curve (see Table 2). Neural networks build at least one hidden layer between the input and output, and the benefits of neural networks to increase model performance actually come from the algorithms' capacity to develop several hidden layers. The training process of neural networks mainly consists of two steps. Firstly, *Feed Forward* takes the inputs or previously hidden layer and combines them with weights. Secondly, *Backward Propagation* takes the output layer or its previous hidden layers to adjust based on the error between the actual and the predicted values. By iteration of this feedforward and backward propagation, neural networks train to adapt the transformation and regression parameters. If no careful process of testing and cross-validation after training is implemented, neural networks are susceptible to overfitting. Regularisation can solve this problem through cross-validation or bootstrapping (Harrell 2001). Another way is to utilize the Bayesian framework. Rather than giving a point estimation, it calculates the distribution of parameters to avoid overfitting problems (Neal, 1996). Moreover, while neural networks tend to be overconfident even when predictions are incorrect and are vulnerable to adversarial attacks (Szegedy et al., 2013), Bayesian neural networks, which produce an ensemble of neural networks, are robust and accurate (Carbone et al., 2020). This may be particularly relevant to increase trust and social acceptance in social and health sciences also in the light of the trade-off in interpretability due to the complexity of the algorithms and resulting models.

Prediction problems are highly relevant in social and health sciences: We may want to predict a certain output, that is, health or social outcome either as accurately or as parsimoniously as possible. Research goals may be to explain maximum variance in the outcome or find a minimal or optimal predictor set. We may want to evaluate how well a certain input, for example, a candidate risk factor, is able to predict an outcome. Prognosis in its simplest form is a prediction problem. There may be defined end points, and we wish to estimate the probability of reaching one of the endpoints. With a perspective of ML as letting the computer/the algorithm define the model instead of the human, ML can test the relative importance of one or more predictors, by considering a large set of covariates, and provide absolute values of importance or rank-order information. Again, the curse of dimensionality mentioned above applies. For in-depth explanations, we refer to prediction textbooks (Kuhn & Johnson, 2013).

A typical problem in the health and social sciences is the prediction of rare outcomes, such as disease, crime, learning difficulties, divorce, etc. where only a very small percentage of the observed population will show the outcome of interest. For example, the rate of offenders is very small compared to the total population; dementia is prevalent in less than 10% of the population aged 60 and older; up to one third of married people will file for divorce. Using ML to predict rare outcomes (and rare may be defined as anything less frequent than 50% of the cases), classification algorithms will usually simply develop a model which will only predict the non-occurrences of the outcome, since the algorithm will detect that a guess of "0" will be correct in most cases. Researchers may solve this problem by redefining the outcome, for example, as regression instead of classification problem. Changing the distribution of the outcome in the sample by oversampling of the group with the rare outcome through, for example, synthetic minority oversampling technique, SMOTE (Chawla et al., 2002), is well established. Equally, undersampling of the group without the outcome is possible. Simulated datasets and 'virtual cohorts' may be useful in some cases. We discuss below transforming a rare outcome in a more frequent one by redefining the research question.

**Estimate and project prevalence of adverse social or health outcomes**
Estimating disease prevalence is the basis for quantifying health burden, the need for interventions, and health and social care planning. Estimation of the rate, incidence, or prevalence of a phenomenon under study, for example, a health outcome such as diabetes in a certain population can be considered descriptive; the projection of estimates would rather be predictive. An example, incorporating the social determinants of health perspective, is to take prevalence estimations of non-communicable diseases by age, sex, and race/ethnicity (can be done with individual-level data or aggregate data), and estimate their prevalence for unknown areas through ML. Presence of six non-communicable diseases (NCD) was estimated through LASSO to predict population-level prevalence of NCD with a minimal demographic dataset for 50 U.S. states (Luo et al., 2015). Similarly, health outcomes have also been predicted on the more fine-grained neighbourhood level (Feng & Jiao, 2021).

**Predict adverse health or social outcomes**
We first discuss research with the main interest in the "output", that is, research with the aim to predict a health or social outcome as accurately as possible. This is done by adding more information (variables) to the model, or by accounting for higher-order interactions or non-linear relationships. Typical research questions have been to test improved predictive accuracy of ML compared to traditional modelling, e.g. regarding the social determinants of health (Seligman et al., 2018), the prediction of dementia (Aschwanden et al., 2020; Weiss et al., 2020), or prediction of student drop-out (Kemper et al., 2020), with ML approaches typically not dramatically outperforming traditional regression-based modeling. A good illustration of this research goal is a retrospective prediction of veteran suicides. Bayesian Additive Regression Trees (BART) were the best ML algorithm of several tested in a dataset of veteran suicides matched with a 1% random matched sample of veterans alive at the time (Kessler et al., 2017). Other research has employed the SuperLearner for mortality risk prediction (Rose, 2013), or penalized linear regressions to predict healthcare costs with penalized linear regressions (Kan et al., 2019). A systematic review on *clinical risk prediction* with machine learning found little benefits of ML methods over regressions, and criticized a number of shortcomings in the literature up to that date, particularly the lack of calibration, i.e. testing the reliability of predictions (Christodoulou et al., 2019). Researchers may be aware that the more *distal* the predictors, the more difficult it will be to arrive at robust, accurate predictions. As an example, a study used characteristics of correctional facilities and aggregate inmate characteristics to predict prison violence, assessed by number of inmate-on-inmate assaults with the SuperLearner but failed to arrive at high levels of accuracy (Baćak & Kennedy, 2019).

Interest in the "output", that is, the health or social outcome, may also come with the aim of finding a minimal or optimal (most parsimonious) predictor set. Coming back to the earlier mentioned example to predict population-level prevalence of non-communicable diseases, these diseases were estimated with a *minimal* socio-demographic predictor set (Luo et al., 2015). In dementia risk prediction, for which until recently no robust algorithms were available (Licher et al., 2018), an *optimal* model predicting dementia over 10 years has been recently developed with LASSO (Licher et al., 2019). Preliminary research has developed a parsimonious model to predict differential diagnosis of dementia (Guest et al., 2020). An optimal predictor set was sought in one study to explain variance in firearm violence, combining LASSO and random forest algorithms (Austrom et al., 2018).

A rare outcome may be more frequent in pre-selected samples, for example, dementia research has partly focused on improving prediction of conversion from Mild Cognitive Impairment (MCI) to dementia. At-risk samples of people with MCI may be used to train a model that discriminates converters from non-converters based on questionnaire-, biomarker- or imaging-based variables. Among numerous examples, some studies tested the role of cognitive reserve to predict conversion to dementia with different ML algorithms (Facal et al., 2019), and tested their model developed with a support vector machine algorithm in new subjects (Grassi et al., 2019). While dementia risk prediction models improve substantially particularly after adding genetic and imaging information to the models, there is no significant progress in the ability to predict decline of cognitive performance tests in at-risk samples (Marinescu et al., 2020).

In clinical practice, the relevant "output" may not be an adverse health outcome, but the necessity to intervene. Prediction of optimal timing of clinical decisions has been researched, for example, by the van der Schaar lab. ML architecture, such as the so-called *Autoprognosis*, has been developed to find the optimal timing for referring patients with terminal respiratory failure for lung transplantation (Alaa & van der Schaar, 2018) and to predict adverse cardiovascular outcomes better than traditional risk scores (Alaa et al., 2019)

**Identify new or evaluate known risk factors**
With an interest in the "input", that is, predictors of a social or health outcome, some research has used large predictor sets to identify previously unknown predictors of a social or health outcome, or to evaluate their predictive ability in the context of the other variables. Aside from the curse of dimensionality that needs to be considered, testing the new predictor set with several ML approaches may be helpful to balance limitations. Studies with this aim have tested, for example, candidate modifiable factors associated with childhood cognitive performance (LeWinn et al., 2020). A study investigated lifestyle factors known to be linked with cognitive functioning, measured by wearables, on their association with cognitive functioning assessed through MMSE scores (Kimura et al., 2019). While no causality could be established due to the cross-sectional nature of the study, visualization through partial dependence plots revealed non-linearities, such as associations plateauing off after a certain threshold, or associations with inverse U-relationships (Kimura et al., 2019), helping to improve our thinking about dose-response relationships in the exposure-outcome associations. A study with a clear observational design estimated the effect of fruit/vegetable density in mothers-to-be on adverse pregnancy outcomes with the SuperLearner (Bodnar et al., 2020). Another study tested the associations of childhood adverse experiences with intelligence in a cross-sectional design (Platt et al., 2018).

**Processes and deviation from normal processes**
Researchers may be interested in moving beyond descriptive research when investigating trajectories of social and health outcomes, and instead adopt a predictive lens if different states or trajectories are already defined by the topic, for example, disease severity or educational or occupational level. A study investigated predictors of chronic obstructive pulmonary disease of highest severity, and common disease trajectories with gradient boosting and a shifting time window approach in health claims data: The authors identified a number of diagnoses (e.g. respiratory failure), medications (e.g. anticholinergic drugs) and procedures associated with a subsequent chronic obstructive pulmonary disease diagnosis of highest severity (Ploner et al., 2020). The temporal patterns detected in this study represent rather order of claims than causal pathways (Ploner et al., 2020), in other contexts, detected temporal patterns may be more robust.

Researchers with an interest in processes of aging may want to define normal aging-related trajectories of health or social outcomes. Then, deviations, defined as a predictive problem, from the normal trajectory could be identified for example with gradient boosting (Er et al., 2017; Anatürk et al., 2020). Researchers should be aware that this goal poses strong requirements on data, as defining the "normal" aging trajectory is not trivial. Ideally, enough information is provided to ensure interpretability and replication.

**Table 2. Overview and non-technical description of most common performance metrics to evaluate ML models**

| Indicator | Explanation, example |
|---|---|
| **Unsupervised learning** | |
| **Specific to clustering** | |
| Adjusted rand index | Measure for the similarity of two data clusterings; related to accuracy but for unlabelled data |
| Mutual information | Measure of mutual dependence, evaluates difference of joint distribution of two sets of variables to the product of their respective marginal distributions |
| Calinski-Harabasz | Criterion to determine the "correct" number of clusters, several related criteria, e.g. implemented in kml |
| Dunn index | Validates clustering solutions |
| **Specific to dimensionality reduction** | |
| Reconstruction error | Measure of the distance (e.g. euclidean for continuous data) between the observed data and the "reconstructed data" from the inferred low-dimensional latent variables |
| **Supervised learning** | |
| Variable importance | Variable importance quantifies the individual contribution of a variable to the classification or regression performance. Several implementations exist. For tree-based models such as random forests, variable importance is often modelled as the sum of improvements gained by using the variable in a split, averaged across all trees (Friedman et al.,2001). In classification and regression, usually the 5-10 most important variables can be meaningfully interpreted. As importance is not easily comparable to traditional statistics metrics, researchers may compare variable importance across multiple models, and add a random variable as benchmark in the consideration of statistical versus clinical (applied) importance |
| **Specific to Regression** | |
| Accuracy | Rate of correctly classified instances over all predictions (= true positives + true negatives/ true positives + true negatives + false positives + false negatives) Good measure in (close to) balanced data, i.e. outcome classes of similar rate |

| | Don't use in imbalanced data |
|---|---|
| Balanced Accuracy | Arithmetic mean of sensitivity and specificity (see below); average of the proportion of correctly classified cases of each class individually, relevant in imbalanced classes |
| Root mean squared error (RMSE) | Average squared difference between target value and predicted value by the model. Penalizes large errors |
| Mean squared error | Preferred metric for regression tasks<br>Average of the square of the difference between original and predicted values |
| Rank order | To evaluate importance of predictors across models or across samples, e.g. to predict dementia (Weiss et al., 2020) |
| Subdistribution Hazard Ratios | To evaluate single predictors in regressions, e.g. to predict dementia (Weiss et al., 2020), can be reported with confidence intervals |
| Mean absolute error | Average of the absolute difference between original and predicted values |
| **Specific to Classification** | |
| Area under the curve (AUC) | For the classification of dichotomous outcomes, this metric specifies the area under the ROC curve, see below, with a range between 0 and 1. The larger value, the better the model. |
| Logarithmic loss/ Log-loss | Measures performance of a classification model, which provides predicted class probabilities. The log loss will get larger when the deviation of the predicted probability from the actual class value increases. Penalises false classification. Works well in multi-class classifications (see multinomial logarithmic loss). Minimising log loss will result in greater accuracy |
| Mean absolute percentage error (MAPE) | Measure of prediction accuracy. It usually expresses the accuracy as a ratio |
| Precision | In classification with two classes, relevant when the cost of false positives is high<br>The proportion of correctly classified cases of all classified cases (e.g. subjects), percentage, = true positives / (true positives + false positives) |
| Percentage correctly classified | Useful for multi-class classification (with more than two categories), easily interpretable |
| Recall or sensitivity or true positive rate | In classification, relevant when the cost of false negatives is high<br>The rate of correctly classified cases of all actual positive cases (true positives by (true positives + false negatives)<br>Correctly classified cases in different strata, e.g. stratum with highest risk, with medium risk, with lowest risk etc. e.g. to predict veteran suicide (Kessler et al., 2017) |
| Receiver operating characteristic (ROC) curve | In classification with two classes.<br>Allows to visualize the trade-off between the true positive rate against the false positive rate. The ROC curve shows the performance of a classification model at all classification thresholds. |
| Specificity | In classification<br>Proportion of correctly classified negatives, the rate of true negatives<br>Equal to 1-false positive rate. |
| F1 Score | Combines both precision and recall, i.e. a good F1 Score would mean both false positives and false negatives are low |
| **Specific to causal inference** | |
| Absolute bias estimate | Sensitivity to unmeasured confounding, in treatSens the estimate of the unmeasured confounder to render the effect of the putative cause to zero ("Coeff. on U" in Dorie et al. 2016) |

| Point estimate, difference in proportions | Effect estimate comparing two treatments, e.g. in BART and other algorithms used for causal inference (see e.g. Keele & Small, 2021) <br> Reported with 95% confidence intervals |
|---|---|
| Adjusted risk difference | Evaluation of effect of candidate cause; linked average treatment effect in TMLE, (e.g. in Bodnar et al 2020) <br> Reported with 95% confidence intervals |

## ML for Causal Inference

Much of what social and health science researchers are after is related to finding causes of a certain feature of the world, or consequences of a certain feature of the world, so, often these disciplines will seek answers to causal questions. We want to identify not only predictors but risk or protective factors, for example, to use in prevention of an adverse health or social outcome. If we don't only want to understand determinants but also intervene, we need an understanding of causal determinants of the disease.

Multiple requirements to the statistical analysis need to be fulfilled before satisfying answers can be found to the question of causality. ML for causal inference requires domain knowledge or, in other words, subject matter expertise. It is vital to select variables wisely according to their position on the directed acyclic graph (DAG) describing the assumed causal relationships between variables (Glymour, 2006; Matthay & Glymour, 2020), which is more and more applied in health research (Tennant et al., 2020). Any statistical analysis aiming at *causal inference* will usually select datasets in which assumptions of causal inference can be assumed to be fulfilled: *exchangeability* (ignorability), i.e. for all who did not receive a particular treatment, the outcome would be the same as for those who did receive the treatment, had they been treated (counterfactual probability of outcome), *positivity*, i.e. all possible values of every level of exposures for every combination of values of exposures and confounders are available or have been assigned in the dataset, and *consistency*, i.e. an individual's potential outcome under their observed exposure history is precisely their observed outcome (Robins et al., 2000). However, even in contexts of (limited) violations of these conditions, some research progress could be made by defining hypothetical interventions or "target trials" if randomized controlled trials (RCTs) are not an option from an ethical perspective, or not feasible for other reasons. Using observational data with a potential-outcomes framework, we can emulate a target trial (García-Albéniz et al., 2017; Hernán & Robins, 2016). A recent study emulated a target trial to test effects of interventions of modifiable factors on BMI rebound in childhood using targeted maximum-likelihood estimation (TMLE) to estimate coefficients (Aris et al., 2021).

In line with Hernán et al. (2019)'s categorization, we first consider here ML for counterfactual prediction, but we will extend their framework by also considering ML for causal discovery, that is, methods, in which the causal structure between variables is learned from the data. Employing the framework of structural causal models has large potential for applications in the social and health sciences, even if largely unexplored today.

## Counterfactual prediction

"Counterfactual prediction is using data to predict certain features of the world as if the world had been different" (Hernan et al., 2019). A useful distinction to arrive at an answer if prediction or counterfactual prediction is sought is to ask if the goal is 'to explain or to predict?' (Shmueli, 2010), although we concede that in some cases we may be able to estimate the magnitude of a causal effect but not to explain it (e.g., in a trial) (Hernán et al., 2019). Causal questions in the potential-outcomes framework (for counterfactual prediction) can be answered with traditional methods e.g. regression and more advanced methods, such as marginal structural models (Hernán et al., 2019). However, over the last years ML for causal inference has grown substantially, and is particularly helpful if embedded in frameworks of causal and statistical inference (Balzer & Petersen, 2021).

For *counterfactual prediction*, ML can be used to analyse large sets of observational data by setting up the data in a way that causal assumptions are met, and then use *predictive* ML approaches to answer causal questions (Blakely et al., 2020). Other ML approaches are well suited to address causal questions if the data are set up properly. Examples are Bayesian Additive Regression Trees (BART) (Chipman et al., 2010; Kapelner & Bleich, 2013; described in more detail in Box 2), TMLE (Schuler & Rose, 2017), and Random Forests (Wager & Athey, 2018). We found few examples of applications of reinforcement learning, which can also be conceptualized as a method to approach causal inference in research contexts where data generation (A/B testing) is possible.

In some cases, the causal structure of (part of) the variables will be known, for example, if the variables are both linked and temporally ordered or follow another causal logic (e.g., researcher manipulation of the independent variable or another exogeneous cause). In the data science fields, one would speak of structured high-dimensional input. In these cases, we can use *graphs for ML* to incorporate this causal knowledge, and extend predictive algorithms such as LASSO or neural networks to, for example, fussed LASSO or convolutional neural networks to reflect the causal structure.

---

*Box 2: Detailed description: Bayesian Additive Regression Trees (BART)*

BART is presented here as an interesting approach to tackle counterfactual prediction as it combines typical methods for causal inference, for example, propensity score weighting to balance the probability of treatment assignment and confounder adjustment to calculate counterfactuals which are used to estimate effects of treatment, so many of the necessary researcher decisions are made explicit. In more detail, BART is a sum-of-trees model predicting outcomes as the sum of a collection of individual regression tree fits and an additive Gaussian error term. Each regression tree iteratively applies splitting rules in order to partition the data into non-overlapping subsets, aiming at minimizing the variance within each subset (Chipman, George, and McCulloch 2010; Hill, Linero, and Murray 2020). As single trees overemphasize interactions and struggle to identify true linear relationships, subsequent trees are fit on the residuals predicted values for identified subsets (Chipman et al. 2010; Hill et al. 2020). To avoid overfitting, BART introduces regularization priors for tree size (i.e. the number of subsets/terminal nodes) and shrinkage (i.e. a factor levelling means in subsets). However, the number of trees remains as a tuning parameter for BART models (Carnegie and Wu 2019). A more detailed description of the underlying Bayesian backfitting algorithm can be found elsewhere (see Chipman et al. 2010). *Implementations* of BART exist for both regression and classification settings. Unlike common tree-based ML approaches like Random Forests or Boosting, regularization priors convey flexible tendencies (e.g. towards small trees) rather than fixed parameters identified by computationally heavy grid searches. Priors are further applicable to high dimensional data and smoothing regression functions (Hill et al. 2020). Performance was shown to compete or exceed common approaches like boosting, neural networks or random forests. However, especially for binary outcomes, cross validating BART models to choose regularization priors is advantageous (Chipman et al. 2010; Dorie et al., 2019 cited by Hill et al. 2020).

Besides computational benefits, BART is applicable to a wide variety of research foci and outcomes (e.g. survival, multinomial logistic regressions) and especially well-equipped for causal inference tasks as modelling complex response surfaces and controlling for confounding does not rely on parametric assumptions (Carnegie and Wu 2019; Hill et al. 2020). Resulting posterior distributions allow to readily estimate individual average treatment or heterogeneous causal effects (Carnegie, Dorie, and Hill 2019; Hill et al. 2020). Moreover, the underlying likelihood framework delivers probabilistic statements about the outcome including credibility intervals whereas identifying and quantifying the effect of individual variables on the outcome is more complicated (Carnegie and Wu 2019). Recent adaptations and implementations of BART further allow modelling and including scores for probabilities of treatment as well as simulating treatment effects in the presence of unobserved confounding. In addition to that procedures to control for the lack of common support are available (see Hill et al. 2020).

**Evaluate potential causes of (adverse) social or health outcomes**
With the aim to assess the effect of candidate causes, research designs may be set up to evaluate the ability of a predictor to *causally influence* an outcome in the context of controlling for confounders. One study tested the ability of adverse childhood experiences to influence mental disorders, and racial/ethnic differences with TMLE, discussing possible violations of the conditions for causal inference (Ahern et al., 2016). Another study tested the effects of fruit/vegetable density in nutrition of mothers-to-be on adverse pregnancy and birth outcomes, showing that TMLE outperformed traditional modelling by finding small effects and giving more precise estimates (Bodnar et al., 2020). Applying a causal perspective even helped to solve the obesity paradox in critically ill patients (Decruyenaere et al., 2020). Using spatial data, one study tested alcohol outlet density and norms on alcohol use disorder with targeted minimum loss-based estimator for rare outcomes (rTMLE; Ahern et al., 2015). Parental and individual socioeconomic determinants of income were tested in comparative perspective with regression trees, along with comparative estimations of country-level inequality of opportunity (Brunori et al., 2018). With the aim to improve medical diagnosis by incorporating counterfactual prediction, associative and counterfactual algorithms were compared in their performance to diagnose patients accurately based on symptom presentation, employing among other techniques causal structural learning (Richens et al., 2020).

**Comparative treatment effectiveness**
Here, researchers may evaluate which of several treatments (e.g. intervention vs. control or care-as-usual) is most effective in changing the health outcome. A selective overview of more clinical applications in health services research can be found in (Rose, 2020). As there are similarities in evaluating the "treatment" also in policy evaluations, we refer to an overview of developments on ML-based estimation of average treatment effects in economics in (Athey, 2018). One study tested the targeted prescription of cognitive-behavioural therapy versus person-centred by estimated therapy outcome had they been assigned to the other treatment (= counterfactual outcome) (Delgadillo & Gonzalez Salas Duhne, 2020). Another study tested the predictive ability of a minimal demographic and symptoms dataset on cognitive behavioural therapy outcomes; the results were better than chance but, in the view of the researchers, far from clinical utility in both more heterogenous and more homogeneous real clinical datasets (Hilbert et al., 2020, 2021).

**Identify heterogeneous treatment effects**
In contexts with randomized treatment assignment, or a temporal research design to assess the effect of a new or changed policy, counterfactual prediction to estimate average treatment effects is straightforward. Additionally, researchers may want to identify and describe subgroups who respond differently to treatment, that is, explore heterogeneous treatment effects. Analyses of heterogeneous treatment effects can give answers to questions particularly prominent in public health research: what works best for whom, and when? For the technical explanations, see literature on estimating treatment effect heterogeneity (Imai & Ratkovic, 2013). Random forests have been developed to detect heterogeneous treatment effects (Wager & Athey, 2018). Model-based recursive partitioning, e.g. mob function in the party package (Zeileis et al., 2010), can also be used for subgroup analysis (Seibold et al. 2016). BART has been employed to estimate heterogeneous treatment effects in survey experiments (Green & Kern, 2012; Künzel et al., 2019).

While reanalysis of clinical trials is tempting to better understand possible heterogeneity in responding to an intervention or medical treatment, still some caution is warranted: Heterogeneous treatment effects to reanalyse failed trials can be considered problematic and a form of p-hacking, as trials available for re-analysis usually have been designed to yield average effects (Shalit, 2020). Generally, we recommend that most of these approaches of using machine learning to identify heterogeneous treatment effects be used for hypothesis generation, with specific subgroup effects verified in an external population.

**Assessing and removing bias**
Finally, a number of interesting ML applications have been developed to *quantify and address potential bias* in analyses aiming at causal inference. In the absence of ignorability (no unmeasured confounders), sensitivity to unmeasured confounding may severely limit the generalizability of the study findings. The treatsens package estimates the magnitude of an unmeasured confounder that would be necessary to nullify the association between a treatment and the outcome, however, domain knowledge is needed in this analysis (Dorie et al., 2016). In contexts with limited (or improperly realized) randomization, unbalanced distributions of covariates may be biasing the findings. Here, BART can assess lack of common support (Hill & Su, 2013), and covariate prioritization versus matching can adjust for differential probability to receive treatment (Keele & Small, 2021). An overview of different papers that employ machine learning for inverse-probability weighting and propensity-score matching is given in (Blakely et al., 2020). Particularly in aging research, we recommend systematically assessing bias coming from selective attrition and competing risk of death, e.g. with random survival forests (Ishwaran et al., 2008) that have been applied in studies on dementia (Weiss et al., 2020).

## Causal discovery or: Causal structural learning

In contrast to the Hernán et al. (2019)'s framework, other approaches suggest that causal inference does not necessarily need counterfactual prediction (Pearl, 2009). Although rarely employed in the health and social sciences, learning causal structure from the data is particularly interesting in contexts where data generation (manipulation of treatment) is possible. Even in settings without possibility for treatment manipulation however, such as with observational data, causal structural learning may elucidate causal research questions: What about putative causes that cannot be manipulated in the sense of randomly assigning the exposure in a (real or hypothetical) intervention? We argue that the absence of a hypothetical intervention should not limit us in estimating a causal effect e.g. of sex, as assessing the effect size of a social factor is prerequisite to improving our understanding of the phenomenon under study, and in the development of targeted interventions (Glymour & Spiegelman, 2017). Such a putative cause can be reconceptualized in a way that allows manipulation or some form of intervention, for example, to manipulate perceived race in job applications (VanderWeele, 2020). Going beyond the potential-outcomes framework, however, we argue that in these cases where a hypothetical intervention is absent, we can capitalize on structural causal learning, i.e. using algorithms that are able to learn (and present) causal structure of variables from the data. In the following, we will first give an overview to structural learning before moving to the more complex task of deriving causal inference from graphical structures, so-called causal structural learning.

### Structural Learning

We start by considering structure learning for *undirected graphs*, that is, learning the conditional independence structure across complex high-dimensional data. The graphical lasso (Friedman et al. (2008); Mazumder and Hastie, 2012) is widely adopted in this setting and is based on an underlying Gaussian assumption (i.e. for continuous variables); software includes glasso R package (Hastie et al., 2014); GraphicalLasso in sklearn (scikit.learn.org, 2021). Various extensions of the Gaussian graphical lasso have been developed, including extensions to simultaneously learn and estimate the *network structure* of variables across groups (e.g. corresponding to distinct subpopulations or data collected under different conditions) and across space and/or time (e.g. longitudinal data), for mixed variables (i.e. measurements on both continuous and discrete variables), and for missing data. This is particularly interesting in contexts where data are sparse, that is, contain many empty cells, and the simultaneous processing of all available data would be computationally extremely costly. In a recent study, Li et al. (2020) used latent Gaussian graphical models for mixed variables, that is, binary, continuous, and count variables to infer symptom associations in verbal autopsies, which may be helpful to arrive at more robust classifications of probable causes of death.

### Causal Structural Learning

Causal Structural Learning extends structure learning by also *inferring the direction of the edges in the graph*. Indeed, constraint-based approaches proceed by first learning an undirected graph, representing the skeleton of the DAG, and then determining orientation. Alternatively, score-based approaches search directly over the space of DAGs and score each graph based on a specified objective function; the massive number of DAGs, which grows super-exponentially with the number of nodes/variables, requires carefully constructed search algorithms and scores. Pearl (2009) provides a thorough technical overview of causality, with recent reviews of causal structural learning provided in Scanagatta et al. (2019) and Heinze-Deml et al. (2018). Constantinou et al. (2021) provide an empirical evaluation and comparison of causal structural learning algorithms under noisy data assumptions. Heinze-Deml et al. (2018) conclude that in the application of causal structural learning, performance relies on appropriateness of underlying assumptions (different models rely on different assumptions); sample size has only a weak influence on performance (varied from a few hundred to 20k); and sparser graphs are easier to estimate. At the time of writing we are not aware of any studies in the health or social sciences that have employed this method.

We concede that applications of these methods will need the on-boarding of social and health researchers by collaborators trained in these methods, as using structural causal learning means also there must be efforts to understand and communicate findings. Employing this method will generate more complex sets of results, for example, an increasing number of possibly competing graphs. With a strong emphasis on the potential-outcomes framework in virtually all social and health sciences disciplines, other methods for causal inference (and their potential benefits) remain widely unknown at this point in time.

## Conclusion

Common research questions in the social and health sciences can be mapped to fitting ML approaches, employing distinctions between ML for description, prediction, and causal inference. Again, we would like to emphasise the need to establish a fluid dialogue between researchers from the social and health sciences and methodologically trained researchers to avoid "rediscovering the wheel", as both sides may ignore knowledge or the state of the art of the other perspective. To conclude, ML approaches have potential to considerably improve empirical analysis in the social and health sciences if thoughtfully applied to relevant problems.